# Recent Trends in the Use of Deep Learning Models for Grammar Error Handling

Mina Naghshnejad[1], Tarun Joshi, and Vijayan N. Nair

Corporate Model Risk, Wells Fargo[2]

## Abstract

Grammar error handling (GEH) is an important topic in natural language processing (NLP). GEH includes both grammar error detection and grammar error correction. Recent advances in computation systems have promoted the use of deep learning (DL) models for NLP problems such as GEH. In this survey we focus on two main DL approaches for GEH: neural machine translation models and editor models. We describe the three main stages of the pipeline for these models: data preparation, training, and inference. Additionally, we discuss different techniques to improve the performance of these models at each stage of the pipeline. We compare the performance of different models and conclude with proposed future directions.

## 1   Introduction

Grammar Error Handling (GEH) is a general term that covers both Grammar Error Detection (GED) and Grammar Error Correction (GEC). The parts of the sentences with errors are identified in GED, while GEC deals with applying specific edits to remedy errors and generate the corrected sentences. The area has attracted the attention of NLP community since the 1980s and has been used in word processors as old as Microsoft Word 96.  With advances in parallel computing and the popularity of neural networks (NNs), techniques for GEH have improved substantially in the last decade. Some papers have focused solely on GED and others just on GEC.

This survey presents an in-depth review of the usage of deep neural networks (DNNs) for both GEC and GED. There are a few general surveys on GEC including (Madi & Al-Khalifa, 2018) and (Manchanda, et al., 2016). However, they do not consider more recent developments that apply DNN models. (Felice, 2016) studied synthetic data generation techniques for  GEC and compared random and probabilistic data augmentation and their impact on performance. (Kiyono, et al., 2019) reviewed data augmentation techniques for GEC including direct noise and back (or reverse) translation.

The aim of this survey is to provide a comprehensive review of the research on GEH using deep learning (DL) models as an end-to-end pipeline. The survey focuses on the proceedings of EMNLP[3] and ACLWEB[4]

---

[1] Corresponding author: Email – Mina.Naghshnejad@wellsfargo.com
[2] The views expressed in the paper are those of the authors and do not represent the views of Wells Fargo.
[3] https://2020.emnlp.org/

[4] https://aclweb.org/conference/

conferences, specifically the submissions for BEA GEC shared task of 2019 (Bryant, et al., 2019). In addition, we examine research in selected industrial and academic groups: Grammarly, Microsoft, Google, Cambridge, MIT, Stanford, CMU, and National University of Singapore. We cover both GEC and GED because these two fields are deeply related; techniques from one can be used in the other to improve the performance. Specifically, the survey

- conducts a comprehensive review on deep learning (DL) approaches for GEC and GED;
- provides all related information to make the survey self-contained; and
- discusses open issues and directions in this research field to improve state-of–the-art performance of GEC and GED models.

## 2   Overview of GEH

Early approaches were rule-based, and the rules were determined using heuristics, statistical analysis, or linguistic knowledge (Naber, 2003). With the popularity of machine learning (ML) models, error-specific ML-based GEC models have been increasingly built (Dahlmeier & Ng, 2012). These models classified tokens in input text as `erratic' (with errors) and `correct' tokens and applied error-specific procedures to correct them. Statistical translation machines (SMTs) have also been used for GEC (Rozovskaya & Roth, 2016) (Junczys-Dowmunt & Grundkiewicz, 2016). Since 2016, sequence to sequence (seq-to-seq) GEC models have been developed and have proven to be very effective (Zheng & Briscoe, 2016).

With the success of natural language generation, machine translation was adapted for GEC. Translation-based approaches were built by considering GEC as a seq-to-seq mapping task (Junczys-Dowmunt, 2018) (Wu, et al., 2016). These approaches learn a model that maps (possibly) ungrammatical sentences to grammatically correct sentences. There are also *editor models* where the sequence of edits for source sentences are predicted and used to generate the correct version of the sentences. We will describe each of these approaches in Section 4.

We first explain key terms that will appear in the rest of the paper. We will use the term "Definition" although these are in fact descriptions of terminology.

*Definition 1.  Parallel corpus*
   This refers to corpora of a source text and its equivalent text in a target language. Parallel texts are used for training SMT and NMT models.

*Definition 2. N-gram*
    A sub-sequence of N tokens (or words) is called an N-gram in NLP. For instance the sentence Weather is nice today has the following 2-grams: {'weather is', 'is nice', 'nice toady'}.

*Definition 3.  Language model (LM)*
   A LM determines a probability distribution over sequences of words. Given a sequence of length *m*, say $(w_1, \dots, w_m)$, the model assigns a probability $P(w_1, \dots, w_m)$ to the whole sequence. Estimating the relative likelihood of different phrases is useful in many NLP tasks, specifically language generation problems. Statistical language models are trained using statistics of co-occurrences of N-grams. Recently, DL based language models (LMs) are becoming popular.

*Definition 4.  Sequence labeling*
   Sequence labeling is a pattern recognition task that algorithmically assigns a categorical label to each member of a sequence of observed values Accuracy is generally improved by making the optimal label for a given element dependent on the choices of nearby elements, using algorithms to choose the globally best set of labels for the entire sequence at once.

*Definition 5. Part of Speech (PoS) tagging*

PoS is a sequence-labelling task that refers to the process of marking up a word in a text (corpus) as corresponding to a particular part of speech (noun, verb, adverb, etc.), based on both its definition and its context.

*Definition 6. Name Entity Recognition (NER)*

NER is also a sequence-labeling task, and it deals with identifying and categorizing key information (entities) in text. It is sometimes referred to as entity chunking, extraction, or identification. An entity can be any word or series of consecutive words that refer to the same thing. Every detected entity is classified into a predetermined entity category.

*Definition 7. Confusion set*

For each location in text, LMs generate the probability of occurrence for each word in vocabulary. However, in many cases, we can use additional information to limit the size of probable words for that location. This smaller set of possible words is called a confusion set. For instance before a word with 'noun' PoS, there could be 'the', 'an', 'a' or 'null' token. In this case, the confusion set has size 4.

## 2.1 Evolution of GEC approaches

Figure 1 provides a chronological ordering of GEC techniques that have been developed since the 1980s. Each of them are described in detail in this section.

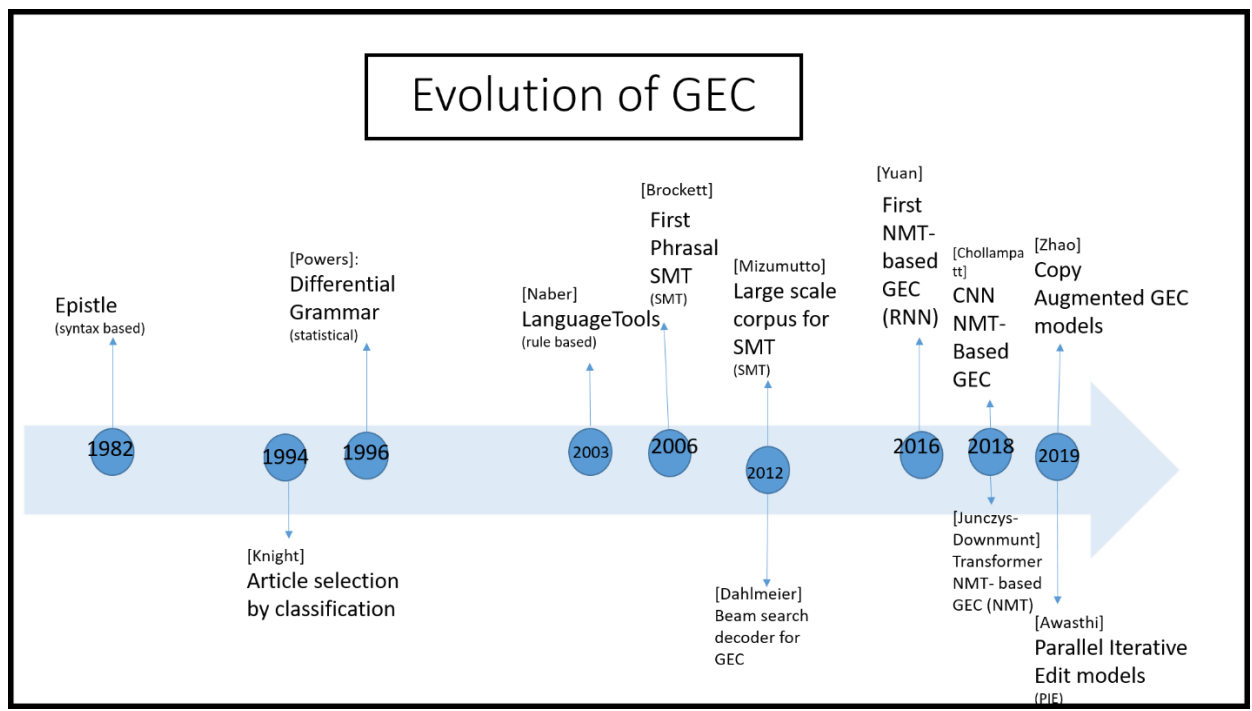

*Figure 1. Evolution of GEC over time*

### 2.1.1 Rule-based grammar error correction

Early approaches for GEC used logical rules, conditioned on features of tokens in the sentences, to determine errors and correct them (Naber, 2003). Some of these rules were generated using parse trees and others were designed heuristically or based on linguistic knowledge or statistical analysis of erratic texts. Rule-based GEC approaches are effective for some types of grammar errors that are more deterministic but they are not able to detect new types of errors.

### 2.1.2 Classifier-based GED models

These were the most popular approaches in early 2000s, and the idea is to train a classifier to detect a specific type of grammar error. Grammar error classifiers take contextualized semantic and syntactic features including PoS tags and NER information and use them to determine the applicable correction. Traditionally, discriminator classifiers like support vector machines (SVMs) and linear classifiers as well as n-gram LM based and Naïve Bayes classifiers were used to predict the potential correction. DNN-based classifier approaches were proposed in 2000s and early 2010s (Rozovskaya & Roth, 2014). However, a specific set of error types have to be defined, and typically only well-defined errors can be addressed with these approaches (Rozovskaya & Roth, 2016).

### 2.1.3 Statistical Machine Translation (SMT) based models

SMT models learn mappings from `source text' to `target text' using a noisy channel model. SMT-based GEC models (Junczys-Dowmunt & Grundkiewicz, 2016) use parallel corpora of erratic text and grammatically correct version of the same text in the same language. Open-source SMT engines are available online and include Moses (Koehn, et al., 2007), Joshua (Li, et al., n.d.) and *cdec* (Dyer, et al., 2010).

### 2.1.4 Neural Machine Translation (NMT) based models

An NMT is an end-to-end model that maps source sentences to target sentences. It is composed of an encoder that maps the input sentences to hidden representation and a decoder that maps hidden representations to text in target language (Wu, et al., 2016). Comparing the output of the SMT baseline with that of the NMT system reveals that NMT captures some learner errors missed by SMT models. One possible reason is that the phrase-based SMT system is trained on surface co-occurrence and therefore unaware of syntactic and linguistic structure. SMT memorizes repeated grammatical patterns from training data. NMT, on the other hand, is able to encode structural patterns from training data and is more likely to capture an unseen error (Grundkiewicz, et al., 2019).

GEC is related to text translation in many aspects. Researchers have looked at GEC as a specific type of machine translation problem: we assume that the text is translated from a second-language English speaker to a text of a fluent English speaker. Similar to translation, the use of LMs appears to be a proper approach for compiling and refining the output text for GEC. However there are some significant differences between GEC and translation. There are not so many corpora for GEC; for MT, there is large amount of bilingual text available. In GEC, many words and structures are kept the same, and there is not as much lexical or positional divergence On the other hand, a second-language learner or a careless writer may make mistakes without any pattern.

Although NMT has emerged as a powerful approach for error correction, there are multiple factors that limit its performance for GEC. First is the scarcity of training data as there are not so many annotated corpora for GEC tasks. Second, most of the original text is not used during grammar correction. For example, in CoNLL2014, only 15% of the text was corrected by annotators (Felice, 2016). Multiple techniques have been proposed to improve the performance of NMT models. We will also review these techniques in this survey.

### 2.1.5 Editor models

A new paradigm in GEH is to predict edits for input tokens rather than predicting corrected tokens. In these models, the encoder is similar to NMT-based models but the decoder outputs edits instead of target tokens. These are called `editor models'. Predicting edits instead of tokens allows the model to pick the output from a smaller confusion set. This will lead to faster training and inference of GEC models (Malmi, et al., 2019).

### 2.1.6 Comparison of the different models

Table 1 provides a comparison of the advantages and disadvantage of the various GEH techniques that we have discussed so far.

*Table 1. Comparison of GEH models.*

| approach | advantages | disadvantages |
| --- | --- | --- |
| Rule based/ syntax based | <ul><li>easy to incorporate domain knowledge of linguistics</li><li>Rules are extendable</li><li>Rules can be easily configured</li><li>Provides detailed feedback on errors</li><li>The error in ungrammatical sentence can be easily identified based on the constraints which are relaxed during parsing</li></ul> | <ul><li>Requires complete list of grammar rules to cover all type of errors and corrections</li><li>Constraint relaxation technique is not well suited for parsing sentences with missing words</li><li>Failure of parsing does not always ensure that the input sentence is ungrammatically wrong</li><li>Robust parsers with sufficient linguistic rules are not available</li><li>Suffer from curse of natural language ambiguities which unnecessary produce more than one parse tree</li></ul> |
| Statistical rule based (corpus derived rules) | <ul><li>Good results when train and test are similar</li><li>No need to deep knowledge of grammar/linguistics</li><li>Language independent system can be developed</li></ul> | <ul><li>Data sparseness introduces challenges</li><li>Most of the times comments and feedbacks are not provided for errors</li><li>Sometimes a correct sentence is predicted as wrong</li><li>Hyper parameters are estimated heuristically and may change from dataset to another(even btw test and train)</li></ul> |
| Classifier based | <ul><li>Effective for specific type of errors (article determination, subj v agreement)</li><li>Can use advancements in classification techniques to build a strong classifier</li></ul> | <ul><li>Cannot employ unlimited number of classifiers to cover all errors</li><li>Each classifier corrects a single word for a specific error category individually. This ignores dependencies between the words in a sentence.</li><li>Needs to assume limited confusion sets for each term. Not applicable for all error types.</li></ul> |
| SMT based | <ul><li>Is not limited to specific error type</li><li>Does not require grammar/linguistic knowledge</li></ul> | <ul><li>Produce locally well-formed phrases with poor overall grammar (unable to process long range dependencies)</li><li>suffers from the paucity of error-annotated training data for grammar correction</li><li>Multiple components that are not easy to deploy</li><li>Hard to constrain to particular error types</li></ul> |
| NMT based | <ul><li>Is not limited to specific error type</li><li>Does not require grammar/linguistic knowledge</li><li>End to end system that can be deployed easily</li><li>Can benefit from most recent advancements in transformers</li><li>Can benefit from transfer learning</li></ul> | <ul><li>Scarcity of annotated corpora</li><li>Deploying models requires a bit of configuration</li><li>Requires additional computation infra structure</li></ul> |
| Editor models | <ul><li>Requires less training data than encoder-decoder</li><li>Creates less redundant corrections (Higher precision)</li></ul> | <ul><li>The limited vocabulary for target test can lower precision.</li></ul> |

## 2.2 Datasets

The main GEC corpora used in recent research are those provided for BEA-GEC shared tasks and are mainly English as a second learner essays. They are:

- LOCNESS corpus: a collection of approximately 400 essays written by native British and American undergraduates on various topics.
- The First Certificate in English (FCE) corpus: This is a subset of the Cambridge Learner Corpus (CLC) that contains 1,244 written answers to FCE exam questions.
- Lang-8 Corpus of Learner English: This is a somewhat clean, English subset of the Lang-8 website. Lang-8 is an online platform for collaborative grammatical correction of essays for language learners.
- The National University of Singapore Corpus of Learner English (NUCLE): This consists of 1,400 essays written by mainly Asian undergraduate students at the National University of Singapore

## 2.3 Performance metrics

Metrics are needed to evaluate the performance of different approaches. As GEC leads to a sequence of corrected sentences, the metric should compare the output sequences with the *reference* target sequences (that we will call the gold standard sequences). However, note that in order to compare each output sentence with the corresponding reference sentence, the words in the two sentences should be aligned; only then should appropriate metrics be calculated.

There is no unique way of aligning output and reference sequences, and there are multiple algorithms available in the field of computer science. Alignment for GEC is simpler and computationally more feasible than MT because most of the words are the same in the output and the reference sequences. In automatic alignment, the system tries to align subsequences by minimizing a suitable distance. Below we present two distance metrics for sentences and then explain Felice distance proposed in (Felice, et al., 2016).

*Definition 8. Levenshtein distance between two sentences*
  This is a string matching metric for measuring the difference between two sequences. Informally, the Levenshtein distance between two sentences is the minimum number of single-word edits (i.e. insertions, deletions, or substitutions) required to change one sentence into the other.

*Definition 9. Damerau-Levenshtein (D-L) between two sentences (DL used as Deep Learning too)*
  The D-L distance between two sentences is the minimum number of operations (consisting of insertions, deletions or substitutions of a single word, or transposition of two adjacent words) required to change one sentence into the other. Unlike the Levenshtein distance, D-L distance considers switching of two adjacent words as a single edit.

For evaluating GEC results, words in the sentence pairs are typically aligned in a way that most of the similarity between the two sentences is captured. The common approach is based on Levenshtein distance, and it considers the alignment that minimizes the number of insertions, deletions and substitution of single word tokens. However, linguistic similarities of the subsequences are not taken into account and hence is not consistent with human intuition. In addition, GEC edits do not necessarily consist of just a single token. For instance, reordering error, such as [only can → can only], or errors involving phrasal verbs, such as [look at → watch], consist of more than one token on at least one side of the edit. For the reordering [$only\ can \rightarrow can\ only$], Levenshtein distance considers the following edits separately: [$only \rightarrow \emptyset$], [$can \rightarrow can$], [$\emptyset \rightarrow only$]. D-L distance, on the other hand, allows transposition of tokens and hence is better suited for GEC tasks.

(Felice, et al., 2016) generalized D-L distance to match reordered subsequences of source and target of arbitrary length to curate it better for GEC task; we call this *Felice alignment*. Additionally, Felice alignment takes into account lemma and PoS to score token mismatches. It calculates substitution cost as sum of lemma difference, PoS difference, and character difference.

Examples of standard Levenshtein and Felice alignments are shown in Table 2 and Table 3. For non-matching words like (propaganda, publicity) and (companys, companies), partial character-based matching is computed. The Levenshtein distance between source and target is 8 while the Felice distance is about 4.6. It can be seen that Felice alignment is more flexible and captures similarity of the two sequences in a way that is more consistent with our intuition.

*Table 2. Alignment of output and reference sentences using Levenshtein distance*

| source | This | wide | spread | propaganda | benefits | only | to | the | companys | . | |
|---|---|---|---|---|---|---|---|---|---|---|---|
| target | This | widespread | publicity | only | benefits | their | companies | . | - | | |
| d | 0 | 1 | 1 | 1 | 0 | 1 | 1 | 1 | 1 | 1 | 8 |

*Table 3. Alignment of source and reference sentences using Felice distance*

| source | This | Wide spread | propaganda | Benefits only | only | to the | the | companys | . | |
|---|---|---|---|---|---|---|---|---|---|---|---|
| target | This | widespread | publicity | Only benefits | their | - | their | companies | . | |
| d | 0 | 1/11 | 9/10 | 1 | 1 | 1 | 2/5 | 2/9 | 0 | ~4.6 |

After alignment of the output and reference sentence, $M^2$ metrics (Dahlmeier & NG, 2012) are calculated by counting matches between the aligned words. These metrics, which are the most widely used for GEC performance, have three components: precision, recall and $F_\beta$-measure (for a given value of $\beta$). They are computed on Levenshtein alignment of phrase-level edits to gold-standard edits annotated by humans. However, $M^2$ metrics have limitations as they do not reward error detection and reward only corrections that are identical to gold standard (reference) sentence. Additionally, they reward only exact matches of hypothesis with reference sentence. This is an issue because, in many cases, reordered sentences have similar meanings and should be considered equivalent.

(Felice and Briscoe, 2015) proposed the I-measure which is a token-level accuracy-based metric that improves upon $M^2$-scores. It combines measures of error detection with error correction. In addition, it rewards correct detection of errors in the sentence even when the corrections do not exactly match with the reference sentence.

Another, more recent, improvement over $M^2$ is the Errant scoring system (Bryant & Mariano, 2017) which scores the alignment of two sentences using Felice alignment explained earlier. In Errant, partial rewards are considered for corrections with the same stem as the word in gold standard. Additionally, the reordering of reference sentence are also rewarded. Errant was used for GEC shared task of BEA 2019 both for automatic alignment and labelling of the sentences (preparing parallel texts) and for evaluating the submitted GEC models. Most of the recent papers (Zhao, et al., 2019) (Awashthi, et al., 2019) (Omelianchuk, et al., 2020) reported GEC performances with both Errant and $M^2$ metrics.

# 3 Deep Learning (DL) Models for GEC

## 3.1 Terminology
First, we provide descriptions of the mostly used terms in neural networks and DL models:

*Definition 10. Neural networks*
   Neural networks (NNs) are computational algorithms based on a network structure that is loosely based on mimicking biological neuronal networks. The network is made up of multiple layers with many nodes in each layer. There is an input layer with nodes corresponding to features or original observations and output layer that yields the final results. Internally, there are hidden layers with nodes where computations occur: each node takes a linear combination of the outputs from nodes in previous layers and applies a non-linear activation function to yield an output, much like the human brain which fires when it encounters sufficient stimuli. The weights in the linear combinations are trained iteratively using backpropagation. The most common NNs have a feedforward structure where the connections among the nodes go only in the forward direction (left-to-right).

*Definition 11. Deep neural networks (DNN) and deep learning (DL)*
   Deep learning is the name used for learning algorithms based on NNs with several hidden layers (Goodfellow, et al., 2016). The corresponding NN is called deep NN (DNN) and the algorithm is called DL. Typically, a DNN is just a feedforward NN with many hidden layers (deep), but there is no widely accepted threshold that indicates a transition from shallow to deep NNs. Recent usage deals with tens or even hundreds of hidden layers. The term DNN is also used to describe more complex network architecture described below.

*Definition 12. Recurrent neural networks:*
   Recurrent NNs (RNNs) make use of sequential information. The term recurrent refers to the fact that they perform the same task for every element of a sequence, with the output being dependent on the previous computations.

*Definition 13. Long-short term memory (LSTM) NNs*
   LSTM is a form of recurrent neural networks that has additional "forget" gates over the simple RNN (Gers, et al., 2000). It calculates the values for the hidden state of interest by taking a combination of three gates: input, forget, and output gates. The gated units enables it to overcome both the vanishing and exploding gradient problems.

*Definition 14. Convolutional neural networks*
   Convolutional neural networks (CNNs) utilize layers with convolving filters that are applied to local features (Kim, 2014). A number of convolutional filters or kernels (typically hundreds) of different widths slide over the entire word embedding matrix. Each kernel extracts a specific pattern of n-gram. A convolution layer is usually followed by an aggregate-pooling strategy which subsamples the input, typically by applying a max operation on each filter.

*Definition 15. Encoder*
   An encoder is an NN that maps inputs (for example text) to hidden representations. It might be an RNN, a CNN, or a transformer.

*Definition 16. Decoder*

A decoder is an NN that maps information (possibly an encoded hidden layer from encoder) back to its original input form (for example text). Decoders are used for text generation in tasks including essay writing, question answering, and chat-bots conversations. Decoder may be an RNN, a CNN or a transformer.

*Definition 17. Auto-encoder*

An auto-encoder is a type of unsupervised NN used to learn efficient data encodings in an unsupervised manner. The aim of an auto encoder is to learn a representation (encoding) for a set of data, typically for dimensionality reduction, by training the network to minimize impact from noise. The auto encoder tries to automatically generate, from the reduced encoding, a representation as close as possible to its original input.

*Definition 18. De-noising auto-encoder*

De-noising auto-encoders are stochastic extensions of the basic auto-encoder. A de-noising auto-encoder tries to encode the information about the input, while undoing the effect of the stochastic noising to input as much as possible.

*Definition 19. Attention mechanism*

One potential problem with the traditional encoder-decoder frameworks is that the encoder is forced to include information which might not be fully relevant to the task at hand. The problem arises also if the input is long or very information-rich and selective encoding is not possible. The attention mechanism attempts to ease the above problems by allowing the decoder to refer back to the input sequence. The decoder of the attention model, is conditioned on a context vector calculated based on the entire input hidden state sequence.

*Definition 20. Transformer*

One of the bottlenecks suffered by RNN is the sequential processing at the encoding step. Additionally, in CNN models the information used from other parts of the sequence is limited to the few filters specified for the model. To address this, (Vaswani, 2017) proposed the Transformer which based models only on attention mechanisms to capture the global relations between input and output. As a result, the overall architecture became more parallelizable and required less time to train. This also led to better results on tasks ranging from translation to parsing.

*Definition 21. Word embedding*

As NNs and DNNs require numerical inputs, for NLP tasks sequences of tokens are mapped to numerical representations. The trivial representation is to perform one hot encoding and get binary vectors with size of vocabulary for each word. However the one hot encoding is very high dimensional and does not contain semantic representation for the words. For this reason, mappings are learned to map one hot encoding of words to a lower dimensional space (typically 50 to 300) while keeping coordinates of semantically similar words close in the new representation.

## 3.2 DL pipeline for GEC models

DL models differ in two areas: i) DNN architecture (types of NN used for encoder and decoder); and ii) decoder output. Another difference among GEC models is the nature of the decoder output: whether they are corrected sentences (NMT-based) or sequences of edits (editor models). In the first group, the model maps sequences of possibly incorrect tokens into sequences of correct tokens. These models were originally proposed for Machine Translation and were adapted to the task of GEC. In the second group, the sequence of possibly incorrect tokens are mapped to a sequence of edits. As noted earlier, these are called editor models. As both NMT-based and editor GEC models have encoder and decoder, we call them both enc-dec models.

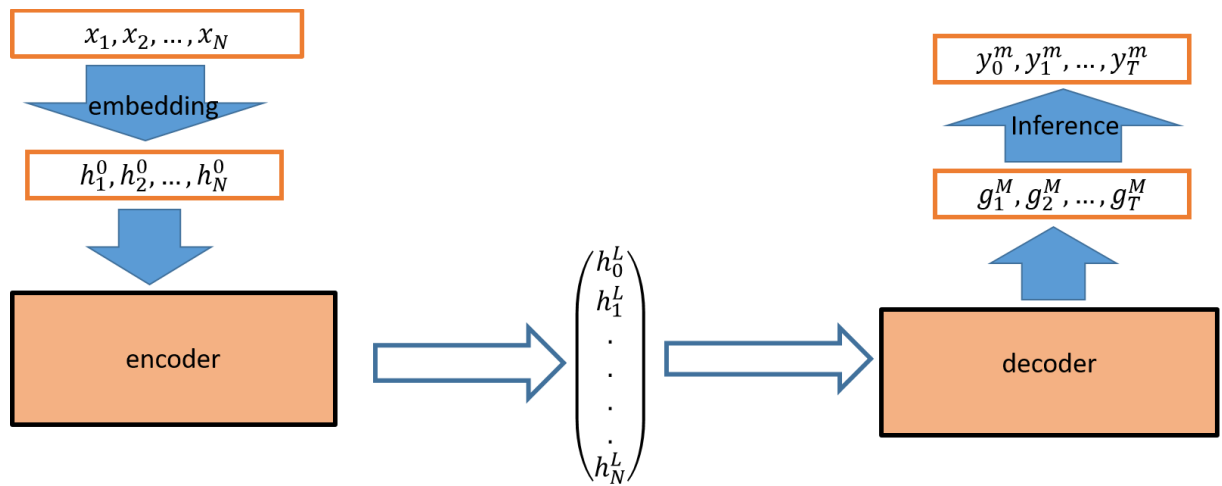

*Figure 2. Seq-to-seq enc-dec architecture*

The goal in all DL-GEC models is to predict the target sequence $(y_1, y_2, ..., y_T)$ from the source word tokens $(x_1, x_2, ..., x_n)$, where $n$ denotes the length of the source sequence and $T$ denotes the length of the target sequence. Figure 2 provides an overview of this process and the underlying enc-dec architecture. The source tokens are mapped into an initial (hidden) layer as $(h_1^0, ..., h_N^0)$. These are further processed through $L$ hidden layers of the NN, and the output from the last hidden layer, $(h_1^L, ..., h_n^L)$, is decoded to get $(g_0^M, ..., g_T^M)$. We consider $m$ layers of decoder in our general formulation. From this, the target sequence $(y_1, y_2, ..., y_T)$ is predicted. This last step is referred to as "inference".

Figure 3 shows the general pipeline of DNN models for GEC tasks. For both NMT-based and editor models, the input sequences are preprocessed and the prepared inputs are used for training a seq-to-seq model. The trained seq-to-seq model is then used to predict the output for new input sequences. It is very common to first pre-train the model weights with a larger synthetic dataset before training with smaller GEC specific parallel texts. In inference step, the proper final output sequence is created. For editor models, inference step includes applying the predicted edits to the input sequence and generating the final corrected sequences. In NMT-based models, the decoder maps the hidden state into sequence of corrected tokens that might or might not be the same size as input sequence. In editor models, the hidden states are mapped into sequence of edits that are of the same length as input sequence.

Decoding is done sequentially. A vanilla decoding includes picking the most probable token at each location in target sequence greedily. During training, using the true tokens at each position of target sequence, the decoder is trained on a next-token prediction loss and outputs the probabilities for words in the vocabulary. Next token is picked based on the element which maximizes the likelihood.

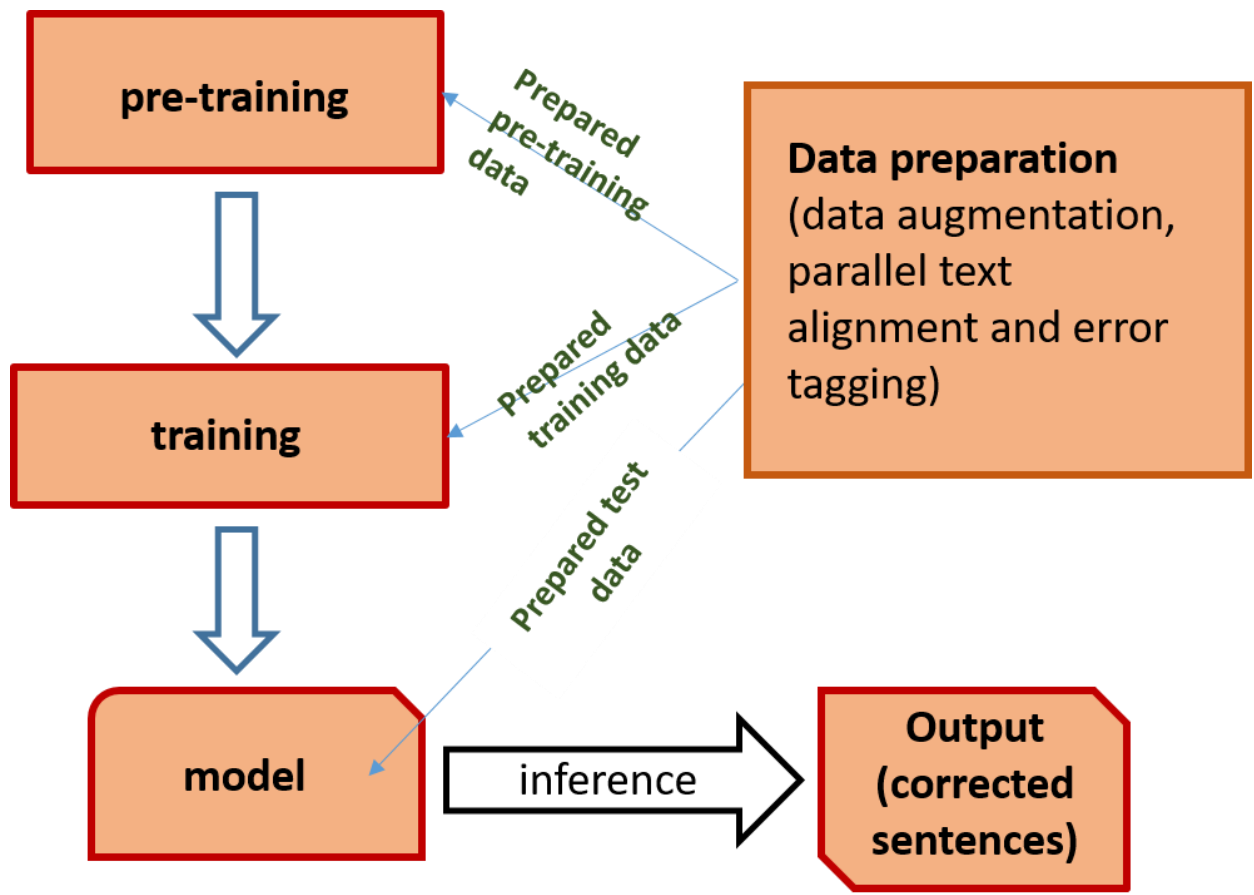

*Figure 3. Seq-to-seq GEC pipeline*

### 3.3 Different NN architectures for GEC: Discussion

Multiple seq-to-seq DL architectures have been proposed for GEC tasks. The first NMT-based GEC, proposed by (Zheng & Briscoe, 2016), used the seq-to-seq model in (Bahdanau, et al., 2014) composed of a bidirectional RNN as an encoder and an attention-based decoder. Multi-layer CNNs were proposed by (Chollapatt & Ng, 2018). Their seq-to-seq GEC model has multiple layers of encoders and decoders. Attention is calculated individually on the output of each decoder. It was shown that LSTM with attention outperforms CNN GEC models in terms of precision while CNNs outperform BiLSTM in terms of recall. CNN seq-to-seq have better ability to capture the context and propose corrections compared to copying the source sentences (Chollapatt & Ng, 2018).

Recently, transformers have gained popularity for many NLP tasks including GEC. About two third of the models submitted to BEA-GEC shared task in 2019 used transformer models (Bryant, et al., 2019). NMT based transformer models (Junczys-Dowmunt, 2018) as well as editor transformer models (Omelianchuk, et al., 2020) have also been proposed. The PIE model (Awashthi, et al., 2019) introduced in Section 2.1.5 is an example of transformer editor models (see Sections 4.2.3 for more discussion on editor models).

### 3.4 Different seq-to-seq models with regard to output

In NMT based models, the inputs, $x_i$s, are potentially ungrammatical tokens and the outputs, $y_i$s, are corrected tokens. In this case, the GEC model is an end-to-end model that outputs corrected sentences by applying the trained seq-to-seq model on input tokens. In editor models, sequence of edits are generated as the output of the model. To produce corrected sentences, an additional procedure of applying edits to input sequences is performed. In essence, editor models can be seen as a type of sequence labelling model. Each token is labelled with the edit that is required for grammar correction. In (Delvin, et al., 2018), token tagging using BERT is applied for the specific task of NER. For token tagging, a token classifier head is used on top of BERT encoder. The confusion set for token tagging classifier is different tags for the editing task. The encoder-classifier model is trained using tagged training data.

# 4 Strategies to Improve Performance in GEC Pipeline

## 4.1 Data and preprocessing

For training GEC models, parallel texts of grammatically erroneous and grammatically correct text are used. These texts are often written by language-learning students and the corrections made by referees. For NMT-based approaches, edit sequences should be generated from parallel sentences to evaluate the result. In Table 4, the reference corrected sentence (ref) as well as the GEC model output is presented along with their corresponding edit sequences (Edit1 and Edit_r). To evaluate GEC performance, these edit sequences are compared.

For editor models, edit sequences are the output of the model as training is done with pair of input sentences and their corresponding edit sequences. The alignment of target and input sentences are discussed in Section 2.3. Another issue is scarcity of parallel data for training. Multiple approaches for data augmentation are presented in Section 4.1.2. Finally, preprocessing of input data for handling out of vocabulary (OOV) tokens is presented at Section 4.1.4.

*Table 4. Edits for the example sentence*

| **src** | This | wide | spread | propaganda | benefits | only | to | the | companys | . |
|---|---|---|---|---|---|---|---|---|---|---|
| **ref** | This | widespread | publicity | only | benefits | | | their | companies | . |
| **output** | This | widespread | propaganda | only | benefits | | | the | Companies | . |
| **Edit1** | M | $S_{widespread}$ | $S_{propaganda}$ | $S_{only}$ | M | D | D | M | $S_{companies}$ | M |
| **Edit_r** | M | $S_{widespread}$ | $S_{publicity}$ | $S_{only}$ | M | D | D | $S_{their}$ | $S_{companies}$ | M |

### 4.1.1 Parallel text alignment and tag generation

All DL approaches perform some sort of seq-to-seq mappings that are used for training and evaluation of the model. For training, final corrected sentences are compared with gold standard reference sentences, and parameters of the model are learned accordingly. For evaluation, the set of edits that map input to the output (corrected sentences) is compared against derived edits from gold standard sentences. Additionally, in many recent works, edit tags are used to guide the correction mechanism. For that purpose, training corpora need to be aligned [and tagged with error types]. As manual alignment and tagging is almost impossible, algorithmic approaches have been used for alignment and tagging of parallel corpora. For BEA 2014 (Ng, et al., 2014) GEC shared task, $M^2$ scoring algorithm was used. For BEA 2019 GEC shared task (Bryant, et al., 2019), Errant scoring was used for alignment of parallel text, and Errant metric was used to measure similarity between the output of submitted models and the gold standard target sentences. These scoring algorithms were reviewed in Section 2.3.

### 4.1.2 Preparing sequences for edit models

To train and test the editor models, we need to generate edit sequences from source-target parallel texts. Edit sequences are extracted from each pair of sentences in the parallel text. Before training the model, edit sequences need to be calculated for pairs of source and target sentences in training set and the vocabulary set is calculated. Table 5 illustrates the edit sequence for two example sentences using (source, target) pairs and calculation of the difference between the pair. The parameters of the DL model are then learned using the labeled training set.

*Table 5. Source and target sequences, their difference and derivation of edit sequence is shown for two examples*

|  | x | y | diff | e |
|---|---|---|---|---|
| **Ex. 1** | [Bolt can have run race] | [Bolt could have run the race] | (C,[)(C,Bolt)(D,can)(I,can, could) (C,have)(I,run,the)(C,race) (C,]) | C C R(could) C A(the) C C |
| **Ex. 2** | [He still won race!] | [However, he still won!] | (C,[)(I,[,However,)(D,He) (I,He,he) (C,still)(C,won)(D, race)(C,!)(C,]) | A(However,)T_case, C, C, D, C, C |

### 4.1.3 Data augmentation

For image analysis, there are natural primitives to introduce noise such as rotations, small translational shifts, and additive Gaussian noise. However, such primitives are not as well developed for text data. Similarly, while de-noising auto-encoders for images have been shown to help with representation learning, analogous methods for text are not well developed. Some recent work has proposed noising— in the form of dropping or replacing individual tokens—as a regularizer while training seq-to-seq models. It has been shown that injecting noise into training data has a smoothing effect on the softmax output distribution (Xie , et al., 2018). Figure 4 shows the common data augmentation techniques.

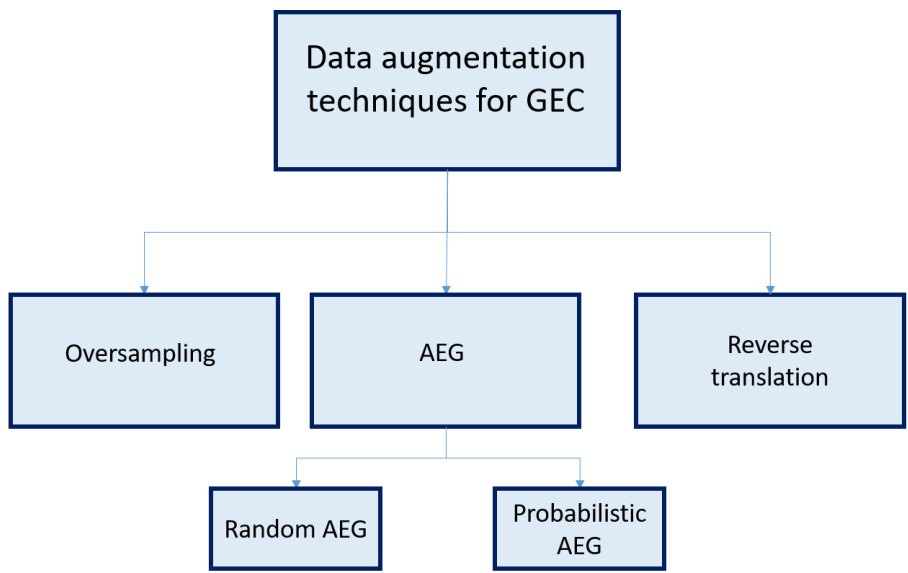

*Figure 4. Data augmentation techniques for GEC*

#### 4.1.3.1 Oversampling

When there is not enough data, (Naplava & Straka, 2019) propose oversampling from the dataset that is most similar to the test set; they add multiple copies of specific subset of inputs to training dataset. Their results show that including multiple duplicates of specific portion in training data improves the GEC performance.

*4.1.3.2 Automatic error generation (AEG)*

Injecting grammar errors into correct sentences, including reordering words, reordering characters, and changing verb tenses, have been used for augmenting parallel texts. The augmentation can be in the form of direct noise, meaning that the errors are randomly injected with some probability into applicable sentences and tokens (Zhao, et al., 2019). A more advanced approach is probabilistic augmentation where one uses the distribution of error types in existing annotated parallel text to select the errors together with lexical or part-of-speech features based on a small context window. Table 6 presents the frequency rate in FCE learners corpora described in Section 2.2.

*Table 6. Examples of error types and their frequency rate in FCE corpora*

| Error type | Example | FCE rate | Error explanation |
| --- | --- | --- | --- |
| DET | It is obvious to see that [internet→the internet] has changed people's lives | 10.86 | Article or Determiner |
| PREP | This article [discuss about→ discuss] whether eating omega-3 is beneficial or not. | 11 | Wrong proposition choice |
| PRON | ours → ourselves | 3.51 | choice of pronoun |
| PUNCT | ! →. | 9.71 | punctuation |
| SPELL | genectic → genetic | 9.59 | spell error |
| VERB | ambulate → walk | 7.01 | choice of verb |
| VERB FORM | to eat → eating | 3.55 | Infinitives, gerunds and participles. |
| VERB TENSE | eats → ate | 6.04 | inflectional and periphrastic tense, modal verbs and passivation |
| NOUN | person → people | 4.57 | choice of nouns |

In a study by (Felice, 2016), the performance ($F_1$ score) of SMT GEC (the state of the art GEC model at the time) on purely artificial grammar error dataset was compared among different error types. The experiments showed that the smaller size of the confusion set leads to more effective artificial data augmentation.

*Table 7. POS tags are used for more realistic error injection into the corpora*

| source | target | PoS | Occurrence probability |
| --- | --- | --- | --- |
| 'play' | 'play' | verb | 0.98 |
| 'plays' | 'play' | verb | 0.02 |
| 'play' | 'play' | noun | 0.84 |
| 'plays' | 'play' | noun | 0.16 |

Using morphological information, such as surrounding words, is known to be useful for more realistic automatic error generation. Incorporating information about PoS tag is often helpful for more realistic error augmentation. For instance, error occurrence probabilities for the word 'play' are shown in Table 7. We can see the error probabilities of this word changes due to its PoS status in the sentence. This information is used in the augmentation procedure to inject errors in a more realistic way. For the example in Table 7, the occurrence of 'plays' instead of 'play' is more probable when 'play' is noun compared to when 'play' is a verb.

*4.1.3.3 Error generation by reverse translation*

In reverse (or back) translation, a model that generates an ungrammatical sentence from a given grammatical sentence is trained. This is also called a reverse model. The output of the reverse model, along with the input as target, are used as pseudo data. The error generation processes use a neural seq-to-seq trained to translate clean examples to their noisy counterparts (Xie , et al., 2018). By training it end-to-end on a large corpus of noisy and clean sentences, the model is able to generate rich, diverse errors that better capture the noise distribution of real data (Xie , et al., 2018).

The vanilla back translation approach is simply reverse noising: a reverse model from $(Y \rightarrow X)$ using the parallel clean-to-noisy corpora trained and standard beam search is used to generate noisy targets $\hat{Y}$ from clean inputs Y. However, this tends to be too conservative due to the "one-to-many" problem where a clean sentence has many possible noisy outputs but they consist mostly of clean phrases. When all these erratic-corrected pairs are input to the system as independent training pairs, they cause the trained model to contain far fewer errors on average than the original noisy text. (Xie , et al., 2018) add noise on a seed corpus of (clean $\rightarrow$ noisy) pairs an appropriate search procedure to get more diversity in the decoded outputs. This solution encourages decoding to stray from greedy solutions.

4.1.4 Handling out-of-vocabulary (OOV) issue and spell correction

One of the challenges of seq-to-seq models is that an erratic text often includes many meaningless words that are possibly misspelled. Spell errors in the training and test data can limit the effectiveness of GEC models. (Zheng & Briscoe, 2016) used an unsupervised word alignment model for SMT and a word-level statistical translation model to replace unknown words in the output. Another approach to tackle OOV problem is to use embeddings on subword vocabularies. Word embeddings like FastText (Bojanowski, et al., 2017) and BERT (Delvin, et al., 2018) are based on subwords. FastText uses bag of character ngrams in each word, while BERT uses Byte Pair Encoding (Rico, et al., 2015). In these approaches, a limited set of subwords are finalized and used to represent words. Using in-vocabulary sub-words for OOV with embedding models can help to retrieve representation for a subset of OOV words and hence improve GEC results. Another, more recent, methodology in handling misspelled OOV is to perform spell correction as a preprocessing procedure. (Zhao, et al., 2019) and (Awashthi, et al., 2019) report that performing spell correction before training end-to-end model improves the GEC task performance.

## 4.2 Training

The training of GEC model involves an optimization process over training samples to learn the model parameters (weight related to encoder and decoder). Starting from an initial set of parameters, the training input output pairs are used to perform backpropagation and update the weights of network.

General DNN architecture for GEC is made of an encoder and a decoder. For NMT-based models the DNN is trained on parallel texts to map ungrammatical sentences to grammatically correct sentences. For editor models, the DNN is trained on input erratic sentences and the sequences of edits. Generally, classification DNNs are trained with calculating maximum likelihood using cross-entropy as the loss function. For seq-to-seq networks, it is common to consider accumulation of cross entropy loss of gold standard sequences.

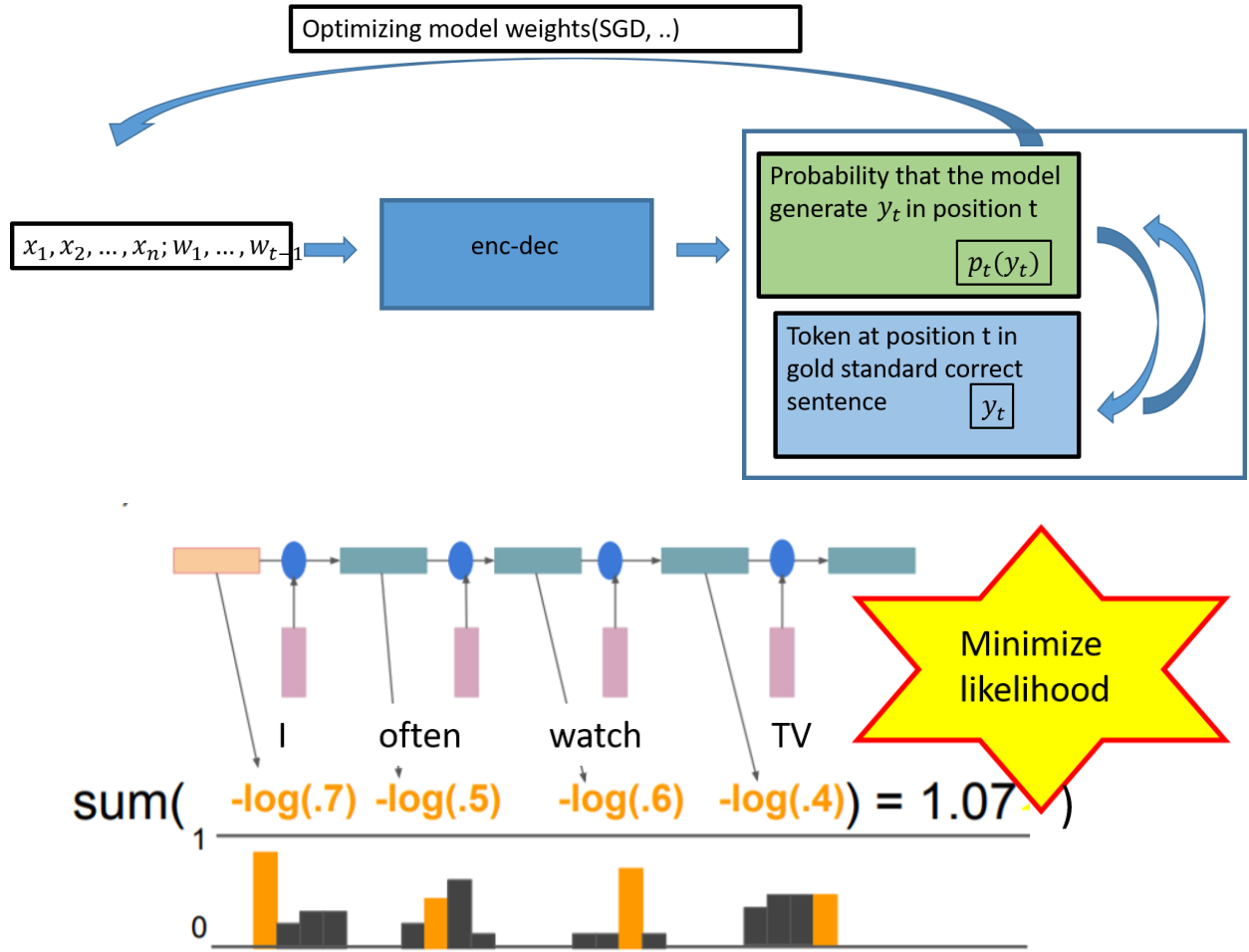

*Figure 5. Top: Training NMT-based GEC model, bottom: an example of training of NMT GEC for a sentence*

As shown in Figure 5, standard maximum-likelihood training for neural machine translation aims at minimizing the negative sum of log probabilities of the ground-truth outputs given the corresponding inputs *(Wu, et al., 2016)* for each sentence. The common loss function that is used to train the network is maximum likelihood of cross entropy:

$$l_{ce} = -\sum_{x_t \in g} \log(p(x_t)), \qquad \qquad Equation\ (1)$$

where $l_{ce}$ is the accumulation of cross entropy loss calculated for each sentence in training set, $g$ is the golden standard output of the model, $x_t$ is the word at position $t$ of the golden standard model. and $p(x)$ is the output probability of the model for token $x$. The cost function is optimized over all pairs of input sentences and golden target sequences in training set.

As this vanilla encoder decoder NN has proven to be sub-optimal (Grundkiewicz, et al., 2019), even with the augmented parallel texts (see Section 4.1.2), several improvements have been proposed for improving the performance of GEC task. We will review these techniques in this section.

### 4.2.1 Transfer learning strategies

First, we describe some additional terminology used in the rest of this section.

***Definition 22. Transfer learning***

Transfer learning (TL) is an ML solution that focuses on using knowledge gained while solving one problem and applying it to a different but related problem (Ruder, 2019). TL is very useful in the context of seq-to-seq DL models because labeled data for a specific task is often scarce. With TL, information from a similar task is learned and used in the current task of interest.

***Definition 23. Pre-training***

Pre-training is an initial training of the network on a large dataset, focusing on training all the parameters of the NN. NNs with pre-trained weights, also called hot start or warm-start, have been shown to outperform cold start networks, where weights are initialized with random or heuristic values. The related data are usually more general and larger corpora like Wikipedia and tasks could be language modeling.

***Definition 24. Fine-tuning***

This is a process to take a model that has been already trained for one task and refine the training (weights) for a new task or on a new data set.

De-noising auto-encoder can also be used to pre-train the network. Here, the pre-training task is similar, but the data is synthetically generated. De-noising auto encoder acts like an NMT. (Zhao, et al., 2019) generated noisy input text by applying random deletion, addition, replacement and shuffling. The encoder-decoder network was pre-trained on de-noising the synthesized erroneous data from one billion word benchmark (Chelba & others, 2014) and then fine-tuned on GEC specific dataset. The weights of this de-noising auto-encoder was then used in the transformer encoder decoder model.

Some other researchers pre-train the network by training an enc-dec network on augmented large scale corpora like Wikipedia. They use direct noise or back-translation to create ungrammatical sentences from Wikipedia corpus and use the resulting parallel corpus to pre-train encoder-decoder NMT model.

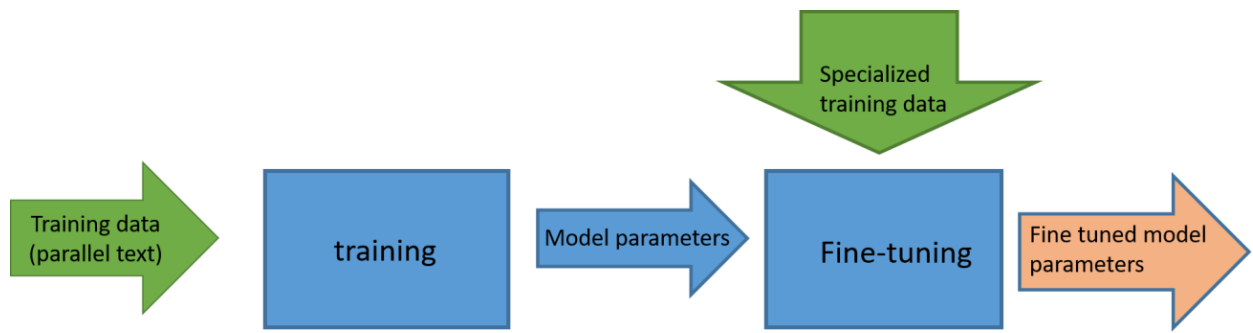

*Figure 6 Transfer learning for GEC models. The model is first pre-trained with larger corpora and then fine-tuned with more related (often scarce) parallel texts*

Bidirectional Encoder Representations from Transformers (BERT) are known to dramatically improve sentence representation learning. BERT pre-trains its encoder using language modeling and by masking random tokens and training the model to guess the masked tokens from surrounding words. Pre-training with BERT allows distributional relations between sentences to be learned in right-to-left and left-to-right direction (Delvin, et al., 2018).

GED has been studied as a sequence labeling task by (Rei, 2017). In their work, an RNN LM is trained along with sequence labelling tasks including PoS tagging, NER and GED. In (Keneko & Mamoru, 2019), BERT model is proposed for labelling tokens as grammatically correct or grammatically incorrect.

### 4.2.2 Copy-augmented seq-to-seq training

One improvement over ordinary NMT-based models is to allow direct copy of input tokens in addition to translating them. Copy-augmentation has also been proposed for abstractive document summarization (Xu, et al., 2020). The architecture allows for direct copying of some source sequences to target sequences. This is intuitively reasonable because many sentences in the target are identical to sentences in source sequences. Copy-augmented transformers are an extension of NMT model with transformer architecture (Zhao, et al., 2019). In these models, a copy attention is calculated in addition to the transformer attention and used to compute the probability of copy for each input token. The final output of a copy augmented transformer has a combination of both generative and copy distributions.

### 4.2.3 Training Editor Models

Editor models can be considered as sequence labelling models. The set of possible labels include all possible edits that can be applied to a grammatically wrong token. They are also called Local Sequence Transduction, as sequences are not mapped to target sequences but edited locally and converted to target sequences.

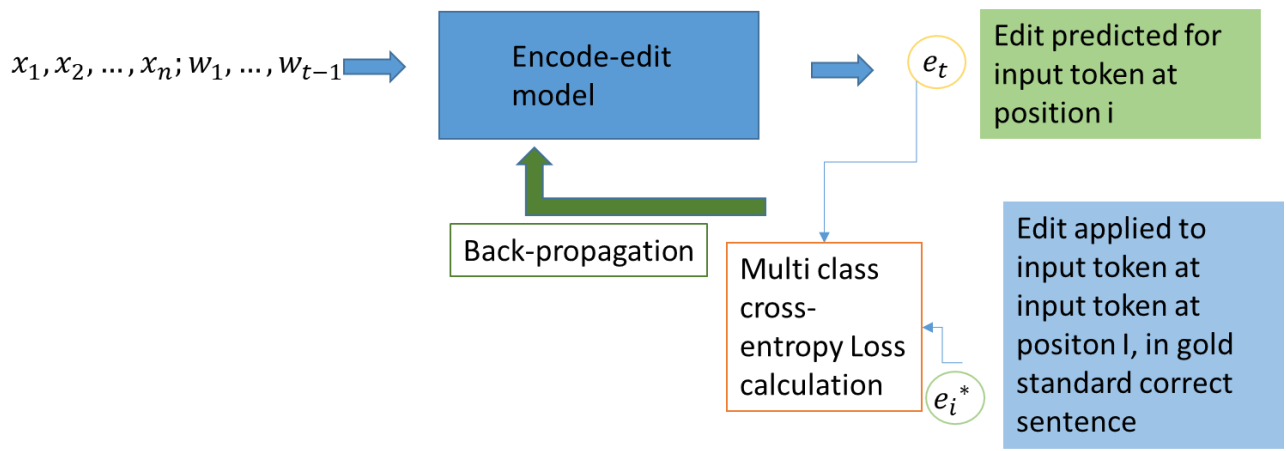

*Figure 7. Training encoder-edit model*

Before training the model, a function called Seq2Edit is used to map input $x$ to a sequence $e$ of edits from edit space $\epsilon$ where $e$ is of the same length $x$ in spite of $x$ and $y$ being of different lengths. The sequence of edits for the gold standard reference sentences are used to train the encoder-edit model as shown in Figure 7.

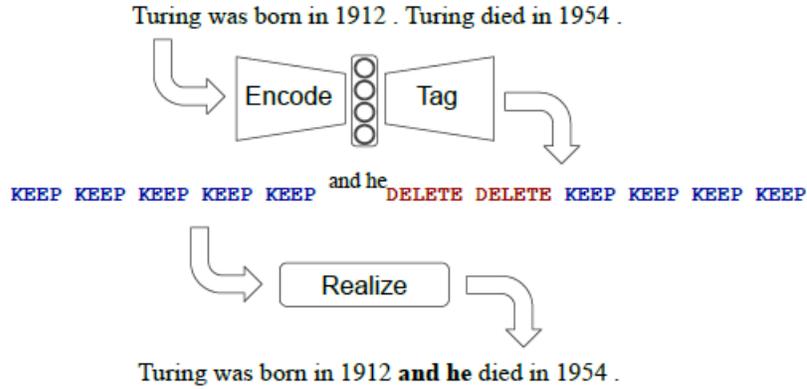

*Figure 8. Laser tagger architecture for sentence editing. Figure from (Malmi, et al., 2019)*

An early editor model, proposed in (Malmi, et al., 2019), is LASERTAGGER which consists of three steps (Figure 8): (i) Encode builds a representation of the input sequence, (ii) Tag assigns edit tags from a pre-computed output vocabulary to the input tokens, and (iii) Realize applies a simple set of rules to convert tags into the output text tokens. In tagging phase, each token is tagged with {KEEP, DELETE}. There is also an added phrase P that can be empty to introduce appending of new tokens to output. The combination of the base tag B and the added phrase P is treated as a single tag and denoted by $^PB$. For encoder, LASERTAG uses BERT base model. For decoder (tagger), Lasertag proposes an autoregressive decoder which is in fact a single layer transformer decoder on top of the BERT encoder.

(Awashthi, et al., 2019) and (Omelianchuk, et al., 2020) have proposed similar encoder-tagger models. (Omelianchuk, et al., 2020) compare usage of different transformer architectures for their encoders and report that XLNeT (Yang, et al., 2019) and RoBERTa (Liu, et al., 2019) outperform pre-training with other transformer architectures. They also decompose the fine-tuning into two stages: in first stage they fine tune on erroneous only sentences and subsequently fine-tune further on a smaller high quality dataset that contains both erroneous and error free sentences.

### 4.2.4 Specialized objective functions

Several GEC specific objective functions have been proposed in the literature to improve performance. (Junczys-Dowmunt, 2018) suggested the use of weighted log-likelihood to avoid converging to a local optimum where the model just copies the input unchanged to the output. The paper proposes assigning higher weights to target tokens that are different from source tokens.

(Wu, et al., 2016) noted that training solely based on cross entropy of words learns only from sentences that are identical to the gold standard and discards information in sentences that are slightly different They showed that incorporating task specific reward function into the loss function improves the performance of seq-to-seq model substantially. See (Wu, et al., 2016) for more details. There are also special objective functions to handle editor models (see (Awashthi, et al., 2019)).

### 4.3 Inference - Generating Corrected Sentences

Once an NMT-based or editor model is trained, it is used to generate corrected sentences from unseen ungrammatical text. For GEC, we have to generate grammatically corrected sentences: the outputs should correspond to input sentences that are grammatically correct. As discussed earlier, for editor models, the output sequences should be generated from the output edits by the decoder. But text generation is a difficult task because the confusion sets of output are large (as large as the complete vocabulary set of the target language). Usually, suitable search techniques are used to improve the quality of the output text by giving the decoder more choices. LMs can also be used to guarantee the fluency of generated texts. Below, we will review GEC-specific solutions including inference in editor models as well as general seq-to-seq strategies including beam-search, iterative decoding, re-ranking with LM, and ensemble decoding.

### 4.3.1 Generating sequences using the decoder output

As explained in Section 3, the seq-to-seq decoder outputs vectors for each token at target sentence. At the inference step, the most probable words in the output sentence are determined by applying the softmax operation below:

$$\boldsymbol{P_t(w) = softmax\ (L^{trg} h_T + b)} \qquad Equation\ (2)$$

Here, $L^{trg}$ is the embedding matrix for target vocabulary and $h_t$ is the final output of the decoder before last layer for token t in target sequence. Figure 9 shows the calculation of probability distribution over the output vocabulary for an output of size $n$. For NMT, the output vocabulary set is the set of vocabulary in the grammatically correct language and for editor models and set of most common grammar edits. The vocabulary size, $v$, is considerably smaller for editor models.

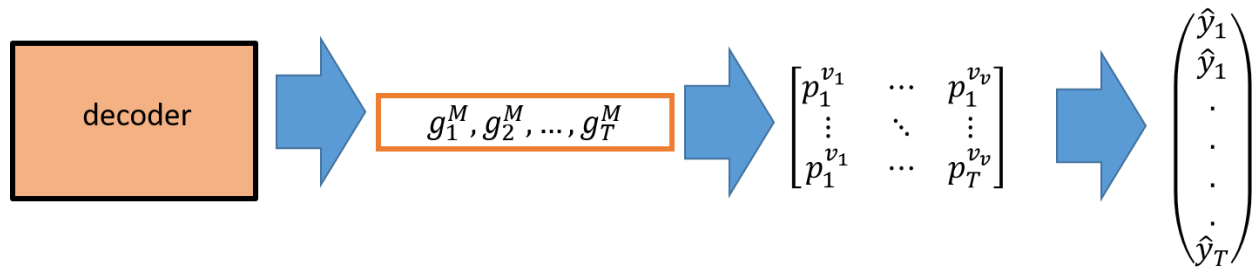

*Figure 9: Inference for seq-to-seq model. $\hat{y}_i$s are predicted corrected sentences in case of NMT model and predicted edits in case of editor model*

Inferring the most likely output sequence involves searching through all the possible output sequences (tokens in the vocabulary) and making the choice based on the likelihood. For this purpose, likelihoods of different sequences are calculated using the likelihood of single tokens in vocabulary. Finally, possible sequences of words are ranked based on sentence likelihood. The size of the vocabulary is often hundreds of thousands of words or even millions of words. Therefore, the full search problem is intractable (NP-complete) and approximations are used.

### 4.3.2 Beam search for inference

Best-first search is a graph search algorithm that outputs all partial solutions (states) according to some heuristic. Beam search is an alternative where only a user-defined number of best partial solutions are kept as candidates. It is an "optimization of best-first search that reduces its memory requirements."[5] Beam search is used in language generation tasks to find the suboptimal target sentence often based on the cross entropy value with the gold standard target sentence for each potential target sentence.

While greedy decoding can give reasonable predicted sequence, a beam search decoder can further boost performance. It does a better exploration of the search space of all possible corrections by keeping around a small set of top candidates as we perform GEC. The size of the beam is called *beam width*; a minimal beam width of, say size 10, is generally sufficient. New sequences are generated by a simple decoder that finds a sequence that approximately maximizes the conditional probability of a trained seq-to-seq model. The beam search strategy generates the GEC output token by token from left-to-right while keeping a fixed number (beam) of active candidates at each time step. As we increase the beam size, performance also increases but at the expense of significantly reducing decoder speed (Freitag & Al-Onaizan, 2017).

---
[5] FOLDOC- computing dictionary https://foldoc.org. Retrieved 8/24/2020.

### 4.3.3 Iterative decoding

In GEC problem, a single sentence may contain multiple errors, so decoding the output sentence in a single run might result in forcing the network to focus on the single most probable correction. Instead, (Lichtarge, et al., 2018) propose applying sequence of corrections on each sentence iteratively. In the models developed by (Awashthi, et al., 2019), the predicted sequence of edits is further refined by iteratively applying the model on the generated outputs to determine additional edits (iterative editing). See Figure 10.

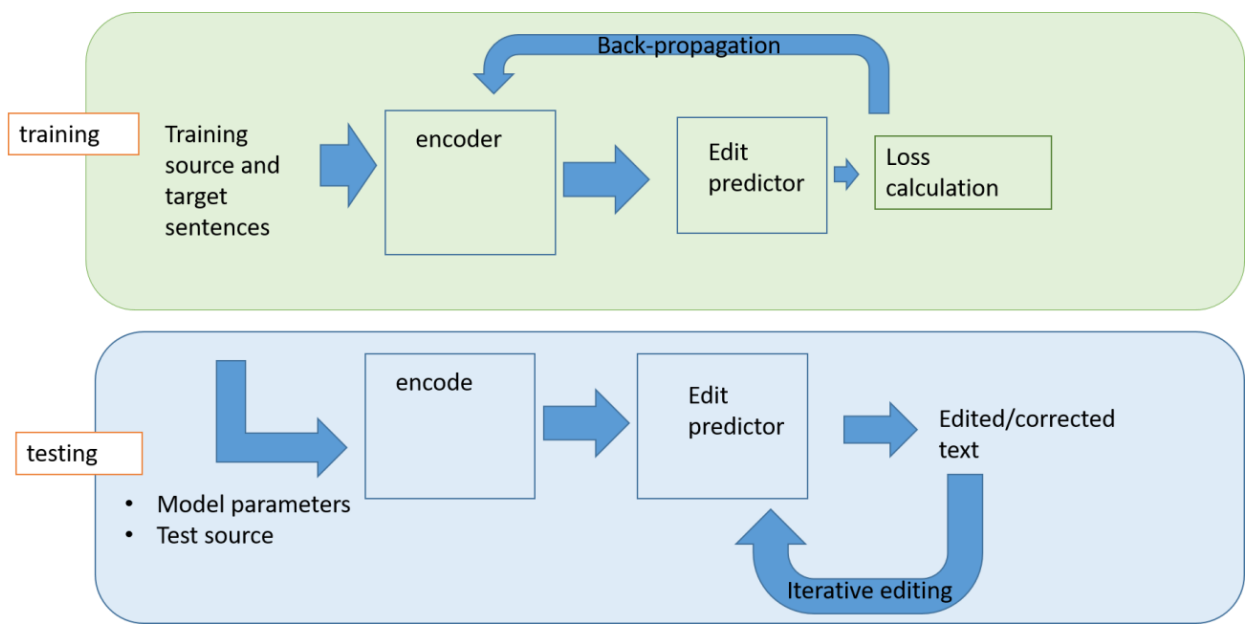

*Figure 10. Parallel Iterative Edit (PIE) structure*

### 4.3.4 Ensemble decoding

Training independent models and "ensembling" the results at inference stage lead to good performance for NMT. For GEC, ensembling improves precision but hurts recall (Junczys-Dowmunt, 2018). The degradation of recall is mainly because the poor recall from different models result in lower recall for the ensemble model. However, ensembling is still helpful because its improvement in precision leads to considerable improvements in some of the $M^2$ scores, and so it is used in many recent GEC applications.

## 4.4 The role of pre-trained LM for GEC task

Using GEC text corpora for training seq-to-seq models introduces limitations for GEH tasks. As the model observes very few changes from source to target, it is hard to learn much semantics from the input parallel source and target text. Pre-trained language models can help fill in this gap and also address the scarcity of training data.

The core idea behind language modelling in GEC is that low probability sequences are more likely to contain grammatical errors than high probability sequences. N-gram language models have been used as feature generators, correction predictors, and prediction rankers for GEH in the past. Recent developments in neural sequence models provided stronger language models (transformers) which use self-attention to capture context in the text. (Bryant & Briscoe, 2018) use 5-gram LM for GEC on a set of grammar error types. Then they use LM to calculate the probability of input sentence and mark the incorrect tokens. LMs generate confusion sets for tokens, and iteratively calculate new sentence probability based on tokens in confusion set and apply the single best correction.

# 5 Discussion of Performance

As discussed in Section 2.3, in order to evaluate GEC models, the outputs of the model are aligned with gold standard correction and precision, recall, and F score are calculated. We need to make sure the same alignment approach is used for calculating the performance metric. $M^2$ is the most common scoring algorithm among mostly cited papers in the last four years, so we use it here for comparison.

*Table 8. Precision and recall comparison for different models. PIE has the highest recall*

| Model | precision | recall | $F_{0.5}$ |
|---|---|---|---|
| **RNN NMT** (Zheng & Briscoe, 2016) | - | - | 39.0 |
| **CNN** (Chollapatt & Ng, 2018) | 65.5 | 33.1 | 54.8 |
| **RNN+Transformer** (Junczys-Dowmunt, 2018) | 66.8 | 34.5 | 56.3 |
| **Copy-augmented Transformer** (Zhao, et al., 2019) | 71.6 | 38.7 | 61.2 |
| **PIE** (Awashthi, et al., 2019) | 68.3 | 43.2 | 61.2 |

Table 8 presents the $M^2$-scores of the most cited GEC models. A baseline NMT-based model that is trained on parallel GEC corpus has $F_{0.5}$ of 39 (Zheng & Briscoe, 2016), (Junczys-Dowmunt, 2018). Using GEC specific edit models improves the recall scores considerably as seen in Table 8. This is probably the result of limiting the confusion sets to specific GEC related vocabulary. The improved recall values is also seen in copy-augmented models. This leads us to conclude that, by limiting the output of GEC, the model learns to detect and correct errors more effectively. Each of the components – incorporating drop-out regularization, handling OOV vocabulary with considering subword embeddings –increases the $F_{0.5}$ for some additional scores.

Another interesting observation is the impact of transfer learning. According to experiments reported in (Zhao, et al., 2019), pre-training decoder weights improved copy-augmented transformer. The PIE editor model has a competitive $F_{0.5}$ with copy-augmented model although it is much faster (Awashthi, et al., 2019). It is important to note that PIE model does parallel encoding and decoding and is much faster than the copy-augmented transformer model. For a discussion of the specific parallel-edit generation algorithm, see (Awashthi, et al., 2019).

A summary of the most important techniques used in data-preparation, training, and inference is presented in Table 9. The Bi-LSTM proposed by (Zheng & Briscoe, 2016) is the vanilla enc-dec model that uses statistical techniques for OOV handling and alignment. Additional techniques (from left-to-right) include using pre-trained models, data augmentation, ensemble decoding, and other techniques discussed in Section 4 to improve performance.

*Table 9 Strategies used by most referenced works on seq-to-seq DL GEC models*

|  | (Zheng & Briscoe, 2016) | (Chollapatt & Ng, 2018) | (Junczys-Dowmunt, 2018) | (Zhao, et al., 2019) | (Awashthi, et al., 2019) |
|---|---|---|---|---|---|
| **Model summary** | Bi-RNN encoder and attention-based LSTM-based decoder | Multi-layer enc-dec CNN NMT-based model | Ensemble of RNN + Transformer | Copy-Augmented transformer +DAE pre-training | Parallel iterative editor model |
| **Data preparation** | ▪ SMT-based OOV handling and text alignment | ▪ Pre-trained word embedding ▪ Subword vocab (BPE) ▪ Train N-gram LM on common crawl (94 B words) | ▪ Data augmentation by oversampling NUCLE corpus ▪ Spell correction as pre-processing | ▪ Training DAE with synthetic parallel corpus ▪ Spell correction as preprocessing | ▪ Pre-training on synthetic data (one billion word corpus) ▪ Spell correction as preprocessing |
| **Training** |  |  | ▪ Dropout regularization ▪ Pretrain decoder by a GRU LM | ▪ Initialize decoder by DAE weights ▪ Multi task learning ▪ Copy-augmented training | ▪ Logit factorized training ▪ Initialize network with pretrained BERT |
| **Inference** | ▪ Beam-search as in NMT | ▪ Use trained LM along edit count features for rescoring, ▪ Ensemble decoding | ▪ Ensemble of RNN and Transformer | ▪ ensemble decoding | ▪ iterative editing ▪ ensemble decoding |

# 6  Conclusions and Future Directions

We have presented a comprehensive survey of DL approaches for GEH. The survey reviewed recent research results and categorized them into two main groups: NMT-based and editor models. We identified three stages of GEC pipeline: data preparation, model training and inference and summarized different techniques for improving performance in each stage. The paper also provides a good set of references to gain insight of the challenges, methods, and issues.

Several algorithms to align sequence pairs have been described and used as a basis to measure performance of GEC models. Another important issue is the impact of transfer learning on performance. In addition, there is a trade-off between output grammatical correctness and fluency. LMs have been exploited to help with generating fluent corrections.

Two promising future directions are apparent. First, incorporating grammar error scoring functions like Errant into training loss function sounds promising. Previously, the usage of GLEU metric (Mutton, et al., 2007) to assess fluency in language generation tasks as a reinforcement learning reward function in the loss function has shown promising improvement (Sakaguchi, et al., 2017). These error scoring algorithms have been used in evaluating GEC models as well as in predicting edits in the inference step. Moreover, the use of parsing trees, for constituency or dependency, (Mirowski & Vlachos, 2015) is worth more attention.

Tree-based grammars are appropriate language models for GEC as the structure of tree will help for detection and correction of errors.

# 7 Appendix: Technical Background

## 7.1 Copy-augmented transformer (Zhao, et al., 2019)

A copy attention layer can be defined as an additional (possibly multi head) attention layer between encoder outputs and the final layer hidden vector at the current encoding step. The attention layer yields two outcomes: the hidden layer output $o_t$ and the corresponding attention scores $s_t$.

$$s_t = softmax\left(\frac{(h_t^{dec})H^{enc}}{\sqrt{d}}\right)$$

$$o_t = H^{enc}s_t$$

The copying distribution is then defined as the attention scores themselves:

$$p^{copy}(y_t|y_{1:t-1};x) = s_t$$

The final output of a copy-augmented transformer has a mixture of both generative and copy distributions. The mixture weight $\alpha_t^{copy}$ is defined at each decoding step as follows:

$$\alpha_t^{copy} = sigmoid(w^\alpha)^T o_t$$

$$p(y_t) = [(1 - \alpha_t^{copy})p^{gen} + \alpha_t^{copy} \cdot p^{copy}](y_t)$$

## 7.2 General BERT based sequence labelling architecture

The output layer of the model will be a softmax over the $i - th$ hidden layer and layer weights to calculate the probability of each edit:

$$\Pr(e_i = e|x) = softmax(w_e^T h_i).$$

Model will be trained using cross entropy loss function:

$$L(e, x) = -\sum_i \log(pr(e_i|x).$$

## 7.3 Factorized logit edit model for PIE (Awashthi, et al., 2019)

Using the BERT transformer model to predict edits, a pre-trained modified BERT LM is trained to learn to predict the edits. During training, two additional attention are calculated: $r_i$ and $a_i$. The outer layer contains edit specific parameters $\theta$. For each position $i$, an additional input of $r_i^0 = [M, p_i]$ where M is embedding for MASK token in the LM. For a potential insert between $i$ and $i + 1$ an additional input comprising of $a_i^0 = [M, \frac{p_i + p_{i+1}}{2}]$ is considered. For layer $l$, self attentions $r_i^l$ and $a_i^l$ are calculated over $h_j^l$ for $j \neq i$ and itself.

The calculation of factorized edits are:

$$logit(e_i|x) = \begin{cases} \theta_C^T h_i + \phi(x_i)^T h_i + 0 & if\ e_i = C \\ \theta_{A(w)}^T h_i + \phi(x_i)^T h_i + \phi(w)^T a_i & if\ e_i = A(w) \\ \theta_{R(w)}^T h_i + 0 + \left(\phi(w) - \phi(x_i)\right)^T r_i & if\ e_i = R(w) \\ \theta_D^T h_i + 0 + 0 & if\ e_i = D \\ \theta_{T_k}^T h_i + \phi(x_i)^T h_i + 0 & if\ e_i = T_k \end{cases},$$

where $\theta_{e_i}^T h_i$ is the score of edit e, $\phi(x_i)^T h_i$ is the score of copying token $x_i$ from source, $\phi(w)a_i$ is the score of appending token $w$, and $\phi(x)$ is the embedding of token x using a pretrained language model. The edit prediction problem is a multi-class classification problem with number of classes equal to $|C| = 1 + |\Sigma_a| + |\Sigma_a| + 1 + |T_k| = 2 + 2M + k$ where M is the size of vocabulary set for appending and replacement ($|\Sigma_a|$) and k is the size of the transformation set ($|T_k|$). **T**raining loss is the sum of cross entropy for each edit $e_i^*$ on each token $x_i$ in gold standard correction. Further, $\theta$ is the set of model parameters including edit specific weights and attention vectors. Here, $\theta = \{a, r, \theta_C, \theta_{A(w)}, \theta_{R(w)}, \theta_D, \theta_{T_k}\}$,

$$l(\theta, x_i, e_i^*) = -\log e_i^\theta(e_i^*|x_i) = -\log\left(\frac{\exp\left(logit\left(e_i^\theta\right)\right)}{\Sigma_{j\in C}\exp\left(logit\left(e_j^\theta\right)\right)}\right) = \Sigma_C P(e_i^* = j|x_i)P(e_i^\theta = j|x),$$

and

$$L(e, x) = -\sum_i \log\left(\Pr\left(e_i^\theta = e_i^*|x_i\right)\right).$$

After optimization, the trained values of edit-specific weights ($\theta_C, \theta_{A(w)}, \theta_{R(w)}, \theta_D, \theta_{T_k}$) and transformer attention, $h_i$ as well as $a_i$ and $r_i$ are calculated and used for prediction. Given an input $x$, the trained model predicts the edit distribution for each input token independent of others:

$$\Pr(e_i|x) = softmax(logit(e_i|x))$$

$$Pr(e|x,\theta) = \prod_{t=1}^n Pr(e_t|x,t,\theta).$$